# A Novel Rough Set Reduct Algorithm for Medical Domain Based on Bee Colony Optimization

N. Suguna[1] and Dr. K. Thanushkodi[2]

**Abstract**— Feature selection refers to the problem of selecting relevant features which produce the most predictive outcome. In particular, feature selection task is involved in datasets containing huge number of features. Rough set theory has been one of the most successful methods used for feature selection. However, this method is still not able to find optimal subsets. This paper proposes a new feature selection method based on Rough set theory hybrid with Bee Colony Optimization (BCO) in an attempt to combat this. This proposed work is applied in the medical domain to find the minimal reducts and experimentally compared with the Quick Reduct, Entropy Based Reduct, and other hybrid Rough Set methods such as Genetic Algorithm (GA), Ant Colony Optimization (ACO) and Particle Swarm Optimization (PSO).

——————————— ◆ ———————————

## 1 INTRODUCTION

THE main goal of feature selection is to find a minimal feature subset from a problem domain with high accuracy in representing the original features [4]. In real world problems feature selection is an important process must due to the abundance of noisy, irrelevant or misleading features. An extensive method may be used for this purpose, it is quite impractical for most datasets. Usually feature selection algorithms involve heuristic or random search strategies in an attempt to avoid this prohibitive complexity. However, the degree of optimality of the final feature subset is often reduced.

Rough set theory [15,17,18] provides a mathematical tool that can be used for both feature selection and knowledge discovery. It helps us to find out the minimal attribute sets called 'reducts' to classify objects without deterioration of classification quality. The idea of reducts has encouraged many researchers in studying the effectiveness of rough set theory in a number of real world domains, including medicine, pharmacology, control systems, fault-diagnosis, text categorization, social sciences, switching circuits, economic/financial prediction, image processing, and so on. However, it is not possible in theory to say whether two attribute values are similar and to what extent they are the same; for example, two close values may only differ as a result of noise, but in the standard RST-based approach they are considered to be as different as two values of a different order of magnitude. Dataset discretization must take place before reduction methods based on crisp rough sets can be applied. This is often still inadequate. However, as the degrees of membership of values to discretised values are not considered at all. To solve this problem, a number of variations in this theory have been proposed. Among these methods, the Swarm Intelligence (SI) based methods perform better than the rest of the methods.

Swarm Intelligence is the property of a system whereby the collective behaviours of simple agents interacting locally with their environment cause coherent functional global patterns to emerge [2]. SI provides a basis with which it is possible to explore collective (or distributed) problem solving without centralized control or the provision of a global model. The SI techniques such as Ant Colony Optimization and Particle Swarm Optimization (PSO) have been hybridized with Rough Set Theory to improve its performance. The major limitations in all these methods are that we have to set more parameter values at random. The performance of the reduct varies based on the parameter settings. Also, the parameters have to be changed based on the applications. In this paper, we have proposed a novel hybridization of rough set theory with Bee Colony Optimization (BCO), which does not require any random parameters. Thus it provides consistent performance, and it is not application specific.

The rest of this paper is structured as follows. Section 2 discusses the fundamentals of Rough Set Theory, in particular focusing on dimensionality reduction; Rough Set Based Attribute Reduction (RSAR), the Entropy-Based Reduction (EBR) and Genetic Algorithm (GenRSAR) based feature selection methods. Section 3 presents the hybrid methods of Rough Set theory with Ant Colony Optimization (AntRSAR) and Particle Swarm Optimization (PSO-RSAR). Section 4 introduces the main concepts of Bee Colony Optimization (BCO) and how this framework can be used for Rough Set Based feature selection. Section 5 details the experimentation carried out and presents the discovered results. The paper concludes with a discussion on the observations and highlights the scope for future work in this area.

## 2 ROUGH SET THEORY

Rough set theory [16] is an extension of conventional set theory that supports approximations in decision making. Rough Set Attribute Reduction (RSAR) [3] provides a filter-based tool by which knowledge may be extracted from a domain in a concise way; retaining the information content whilst reducing the amount of knowledge involved. Central to RSAR is the concept of indiscernibility. Let I = (U,A) be an

————————————————

[1]Assistant Professor in Computer Science and Engg., Akshaya College of Engineering and Technology, Coimbatore, Tamil Nadu, India.
[2]Director, Akshaya College of Engineering and Technology, Coimbatore, Tamil Nadu, India.

information system, where U is a non-empty set of finite objects (the universe) and A is a non-empty finite set of attributes such that $a : U \rightarrow V_a$ for every $a \in A$. With any $P \subseteq A$ there is an associated equivalence relation IND(P):

$$IND(P) = \{ (x,y), \in U^2 \mid \forall a \in P\ a(x) = a(y) \} \quad (1)$$

The partition of U, generated by IND(P) is denoted U/P and can be calculated as follows:

$$U/P = \otimes \{ a \in P : U / IND(\{a\}) \}, \quad (2)$$

Where

$$A \otimes B = \{ X \cap Y : \forall X \in A, \forall Y \in B, X \cap Y \neq \emptyset \} \quad (3)$$

If $(x; y) \in IND(P)$, then x and y are indiscernible by attributes from P. The equivalence classes of the P-indiscernibility relation are denoted $[x]_P$.

Let $X \subseteq U$, the P-lower approximation $\underline{P}X$ and upper approximation $\overline{P}X$ of set X be defined as:

$$\underline{P}X = \{ x \mid [x]_P \subseteq X \} \quad (4)$$

$$\overline{P}X = \{ \{ x \mid [x]_P \cap X \neq \emptyset \} \quad (5)$$

Let P and Q be equivalence relations over U, then the positive, negative and boundary regions can be defined as:

$$POS_P(Q) = \bigcup_{X \in U/Q} \underline{P}X \quad (6)$$

$$NEG_P(Q) = U - \bigcup_{X \in U/Q} \overline{P}X \quad (7)$$

$$BND_P(Q) = \bigcup_{X \in U/Q} \overline{P}X - \bigcup_{X \in U/Q} \underline{P}X \quad (8)$$

The positive region contains all objects of U that can be classified to classes of U/Q using the knowledge in attributes P.

An important issue in data analysis is discovering dependencies between attributes. Intuitively, a set of attributes Q depends totally on a set of attributes P, denoted $P \Rightarrow Q$, if all attribute values from Q are uniquely determined by values of attributes from P. If there exists a functional dependency between values of Q and P, then Q depends totally on P. Dependency can be defined in the following way:

For $P, Q \subset A$, it is said that Q depends on P in a degree k ($0 \leq k \leq 1$), denoted $P \Rightarrow_k Q$, if

$$k = \gamma_P(Q) = \frac{|POS_P(Q)|}{|U|} \quad (9)$$

If k = 1, Q depends totally on P, if 0 < k < 1 Q depends partially (in a degree k) on P, and if k = 0 then Q does not depend on P. Based on this fundamental, here we have discussed two important reduction methods: QuickReduct and Entropy-based method.

### 2.1 QuickReduct

The reduction of attributes is achieved by comparing equivalence relations generated by sets of attributes. Attributes are removed so that the reduced set provides the same quality of classification as the original. A reduct is defined as a subset R of the conditional attribute set C such that $\gamma_R(D) = \gamma_C(D)$. A given dataset may have many attribute reduct sets, so the set R of all reducts is defined as:

$$R = \{ X : X \subseteq C, \gamma_R(D) = \gamma_C(D) \} \quad (10)$$

The intersection of all the sets in R is called the core, the elements of which are those attributes that cannot be eliminated without introducing more contradictions to the dataset. In RSAR, a reduct with minimum cardinality is searched for; in other words an attempt is made to locate a single element of the minimal reduct set $R_{min} \subseteq R$ :

$$R_{min} = \{ X : X \in R, \forall Y \in R, |X| \leq |Y| \} \quad (11)$$

The problem of finding a minimal reduct of an information system has been the subject of much research [1]. The most basic solution to locating such a reduct is to simply generate all possible reducts and choose any with minimal cardinality. Obviously, this is an expensive solution to the problem and is only practical for very simple datasets. Most of the time only one minimal reduct is required, so all the calculations involved in discovering the rest are pointless. To improve the performance of the above method, an element of pruning can be introduced. By noting the cardinality of any pre-discovered reducts, the current possible reduct can be ignored if it contains more elements. However, a better approach is needed - one that will avoid wasted computational effort. The QuickReduct algorithm given in figure 1, attempts to calculate a minimal reduct without exhaustively generating all possible subsets. It starts of with an empty set and adds in turn, one at a time, those attributes that result in the greatest increase in dependency, until this produces its maximum possible value for the dataset.

---

QUICKREDUCT (C,D)
C, the set of all conditional features;
D, the set of decision features.
(1)     $R \leftarrow \{\ \}$
(2)     do
(3)         $T \leftarrow R$
(4)         $\forall x \in (C - R)$
(5)             if $\gamma_{R \cup \{x\}}(D) > \gamma_T(D)$
(6)                 $T \leftarrow R \cup \{ x \}$
(7)         $R \leftarrow T$
(8)     until $\gamma_R(D) == \gamma_C(D)$
(9)     return R

---

Figure 1 The QuickReduct Algorithm

Note that an intuitive understanding of QuickReduct implies that, for a dimensionality of n, $(n^2+n)/2$ evaluations of the dependency function may be performed for the worst-case dataset. According to the QuickReduct algorithm, the dependency of each attribute is calculated, and the best candidate chosen. The next best feature is added until the dependency of the reduct candidate equals the consistency of the dataset (1 if the dataset is consistent). This process, however, is not guaranteed to find a minimal reduct. Using the dependency function to discriminate between candidates may lead the search down a non-minimal path. It is impossible to predict which combinations of attributes will lead to an optimal reduct based on changes in dependency with the addition or deletion of single attributes. It does result in a close-to-minimal reduct, though, which is still useful in greatly reducing dataset dimensionality.

### 2.2 Entropy based Feature Reduction

Another technique for discovering rough set reducts is entropy-based reduction (EBR), developed from work carried out in [6] and is based on the entropy heuristic employed by machine learning techniques such as C4.5 [19]. The motivation behind this approach is the observation that when the rough set dependency measure is maximized for a given subset, the entropy is minimized. For consistent datasets, the resulting entropy is 0 when the dependency degree is 1. EBR is concerned with examining a dataset and determining those attributes that provide the most gain in information. The entropy of attribute A (which can take values $a_1...a_m$) with respect to the conclusion C (of possible values $c_1...c_n$) is defined as:

$$E(A) = -\sum_{j=1}^{m} p(a_j) \sum_{i=1}^{n} p(c_i | a_j) \log_2 p(c_i | a_j) \quad (12)$$

This can be extended to dealing with sub sets of attributes instead of individual attributes only. Using this entropy measure, the algorithm used in RSAR can be modified to that shown in figure 2. Upon each iteration, the subset with the lowest resulting entropy is chosen. This algorithm requires no thresholds in order to function - the search for the best feature subset is stopped when the resulting subset entropy is equal to the entropy of the full set of conditional attributes. However, the entropy measure is a more costly operation than that of dependency evaluation which may be an important factor when processing large datasets.

```
EBR (C)
C, the set of all conditional features;
(1)     R ← { }
(2)     do
(3)         T ← R
(4)         ∀x ∈ (C – R)
(5)         if E(R ∪ {x}) < E(T)
(6)             T ← R U {x}
(7)         R ← T
(8)     until E(R) == E(C)
(9)     return R
```

Figure 2 The Entropy-based Reduct Algorithm

## 2.3 Genetic Based Reduct (GenRSAR)

Genetic Algorithms (GAs) are generally quite effective for rapid search of large, nonlinear and poorly understood spaces [5]. Unlike classical feature selection strategies where one solution is optimized, a population of solutions can be modified at the same time [8]. This can result in several optimal (or close-to-optimal) feature subsets as output.

A feature subset is typically represented by a binary string with length equal to the number of features present in the dataset. A zero or one in the jth position in the chromosome denotes the absence or presence of the jth feature in this particular subset. An initial population of chromosomes is created; the size of the population and how they are created are important issues. From this pool of feature subsets, the typical genetic operators (crossover and mutation) are applied. Again, the choice of which types of crossover and mutation used must be carefully considered, as well as their probabilities of application. This generates a new feature subset pool which may be evaluated in two different ways. If a filter approach is adopted, the fitness of individuals is calculated using a suitable criterion function. This function evaluates the goodness of a feature subset; a larger value indicates a better subset.

The initial population consists of 100 randomly generated feature subsets, the probability of mutation and crossover set to 0.4 and 0.6 respectively, and the number of generations is set to 100. The fitness function is defined as follows (Jensen and Shen, 2003):

$$fitness(R) = \gamma_R(D) * \frac{|C| - |R|}{|C|} \quad (13)$$

## 3. SWARM INTELLIGENCE BASED REDUCT ALGORITHMS

In this section, we have discussed two different the swarm intelligence based reduct algorithms AntRSAR and PSO-RSAR hybrid with Rough set theory.

## 3.1 Ant Colony Based Reduct (AntRSAR)

The ability of real ants to find shortest routes is mainly due to their depositing of pheromone as they travel; each ant probabilistically prefers to follow a direction rich in this chemical. The feature selection task may be reformulated into an ACO-suitable problem. ACO requires a problem to be represented as a graph; where nodes represent features, with the edges between them denoting the choice of the next feature [7]. The search for the optimal feature subset is then an ant traversal through the graph, where a minimum number of nodes are visited, that satisfies the traversal stopping criterion. The heuristic desirability of traversal and edge pheromone levels are combined to form the so-called probabilistic transition rule, denoting the probability of an ant at feature i choosing to travel to feature j at time t:

$$p_{ij}^{k}(t) = \frac{[\tau_{ij}(t)]^{\alpha} \cdot [\eta_{ij}]^{\beta}}{\sum_{l \in J} [\tau_{ij}(t)]^{\alpha} \cdot [\eta_{ij}]^{\beta}} \quad (14)$$

where k is the number of ants, $J_i^k$ the set of ant k's unvisited features, η$_{ij}$ is the heuristic desirability of choosing feature j when at feature i and τ$_{ij}$(t) is the amount of virtual pheromone on edge (i; j). The choice of α and β is determined experimentally.

The overall process of ACO feature selection begins by generating a number of ants, k, which are then placed randomly on the graph (i.e. each ant starts with one random feature). Alternatively, the number of ants to place on the graph may be set equal to the number of features within the data; each ant starts its path construction at a different feature. From these initial positions, they traverse through the edges probabilistically until a traversal stopping criterion is satisfied. The resulting subsets are gathered and then evaluated. If an optimal subset has been found or the algorithm has executed for a certain number of times, then the process halts and outputs the best feature subset encountered. If neither condition holds, then the pheromone is updated, a new set of ants are created and the process iterates once more. To tailor this mechanism to find rough set reducts, it is necessary to use the dependency measure as the stopping criterion. This means that an ant will stop building its feature subset when the dependency of the subset reaches the maximum for the dataset (the value 1 for consistent datasets). The dependency function may also be chosen as the heuristic desirability measure, but this is not necessary.

## 3.2 Particle Swarm Based Reduct (PSO-RSAR)

Given a decision table T = (U,C,D,V,f), the set of condition attributes, C, consists of m attributes. The search space is defined of m dimensions for the reduction problem. Accordingly, each particle's position is represented as a binary string of length m. Each dimension of the particle's position maps one condition attribute. The domain for each dimension is limited to 0 or 1. The value '1' means the corresponding attribute is selected while '0' means not selected. Each position can be "decoded" to a potential reduction solution, a subset of C. The particle's position is a series of priority levels of the attributes. The sequence of the attribute will not be changed during the iteration. But after updating the velocity and position of the particles, the particle's position may appear as real values such as 0.4, etc. It is meaningless for the reduction. Therefore, we introduce a discrete particle swarm optimization for this combinatorial problem [14].

During the search procedure, each individual is evaluated using the fitness. According to the definition of rough set reduct, the reduction solution must ensure that the decision ability is the same as the primary decision table and the number of attributes in the feasible solution is kept as low as possible. In this algorithm, we first evaluate whether the potential reduction solution satisfies POSE = $U_{pos}$ or not (E is the subset of attributes represented by the potential reduction solution). If it is a feasible solution, we calculate the number of 1's in it. The solution with the lowest number of 1's is selected. For the particle swarm, the lesser the number of 1's in its position, the better the fitness of the individual. POSE = $U_{pos}$ is used as the criterion of the solution validity.As a summary, the particle swarm model consists of a swarm of particles, which are initialized with a population of random candidate solutions. They move iteratively through the d-dimension problem space to search the new solutions, where the fitness $f$ can be measured by calculating the number of condition attributes in the potential reduction solution. Each particle has a position represented by a position-vector $p_i$ (i is the index of the particle), and a velocity represented by a velocity-vector $v_i$. Each particle remembers its own best position so far in a vector $Bp_i$, and its j-th dimensional value is $bp_{ij}$. The best position-vector among the swarm so far is then stored in a vector $Gp$, and its j-th dimensional value is $gp_j$. When the particle moves in a state space restricted to zero and one on each dimension, the change of probability with time steps is defined as follows:

$P(p_{ij}(t) = 1) = f(p_{ij}(t-1), v_{ij}(t-1), Bp_{ij}(t-1), Gp_j(t-1))$ (15)

Where the probability function is

$$sig(v_{ij}(t)) = \frac{1}{1+e^{-v_{ij}(t)}} \qquad (16)$$

At each time step, each particle updates its velocity and moves to a new position according to the following equation

$v_{ij}(t) = w \cdot v_{ij}(t-1) + \phi_1 r_1(Bp(t-1) - p_{ij}(t-1))$
$+ \phi_2 r_2(Gp(t-1) - p_{ij}(t-1))$ (17)

$$p_{ij}(t) = \begin{cases} 1 & if\ \rho < sig(v_{ij}(t)) \\ 0 & otherwise \end{cases} \qquad (18)$$

Where $\phi_1$ is a positive constant, called as coefficient of the self-recognition component, $\phi_2$ is a positive constant, called as coefficient of the social component. $r_1$ and $r_2$ are random numbers in the interval [0,1]. The variable w is called the inertia factor, whose value is typically setup to vary linearly from 1 to near 0 during the iterated processing. ρ is a random number in the closed interval [0, 1]. In this step, a particle decides where to move next, considering its current state, its own experience, which is the memory of its best past position, and the experience of its most successful particle in the swarm. The main limitations in the genetic and swarm intelligence methods are: (i) they deal with more parameters in random, the changes made in these values will affect the total performance and these parameters have to be well tuned to deduct better reduct, (ii) these parameter values have to be adjusted based on the dataset (application). To combat this, we have used an artificial bee colony algorithm, which does not require any random parameters to find the optimum reduct as discussed in the following section.

### 3.3 Bee Colony Based Reduct (BeeRSAR)

Nature is inspiring researchers to develop models for solving their problems. Optimization is a field in which these models are frequently developed and applied. Genetic algorithm simulating natural selection and genetic operators, Particle Swarm Optimization algorithm simulating flock of birds and school of fishes, Artificial Immune System simulating the cell masses of immune system, ACO algorithm simulating foraging behaviour of ants and Artificial Bee Colony algorithm simulating foraging behaviour of honeybees are typical examples of nature inspired optimization algorithms.

Artificial Bee Colony (ABC) algorithm, proposed by Karaboga (2005) for real parameter optimization, is a recently introduced optimization algorithm and simulates the foraging behaviour of bee colony for unconstrained optimization problems [10-13]. For solving constrained optimization problems, a constraint handling method was incorporated with the algorithm.In a real bee colony, there are some tasks performed by specialized individuals. These specialized bees try to maximize the nectar amount stored in the hive by performing efficient division of labour and self-organization. The minimal model of swarm-intelligent forage selection in a honey bee colony, that ABC algorithm adopts, consists of three kinds of bees: employed bees, onlooker bees, and scout bees. Half of the colony comprises employed bees and the other half includes the onlooker bees. Employed bees are responsible for exploiting the nectar sources explored before and giving information to the other waiting bees (onlooker bees) in the hive about the quality of the food source site which they are exploiting. Onlooker bees wait in the hive and decide a food source to exploit depending on the information shared by the employed bees. Scouts randomly search the environment in order to find a new food source depending on an internal motivation or possible external clues or randomly. Main steps of the ABC algorithm simulating these behaviours are given in the figure:

(1) Initialize the food source positions.
(2) Each employed bee produces a new food source in her food source site and exploits the better source.
(3) Each onlooker bee selects a source depending on the quality of her solution, produces a new food source in selected source site and exploits the better source.
(4) Determine the source to be abandoned and allocate its employed bee as scout for searching new food sources.
(5) Memorize the best food source found so far.
(6) Repeat steps 2-5 until the stopping criterion is met.

Figure 3 Bee Colony Optimization Algorithms

The above procedure can be implemented for feature reduction. Let the bees select the feature subsets at random and calculate their fitness and find the best one at each iteration. This procedure is repeated for number of iterations to find the optimal subset.In first step of the algorithm, the employed bees produce the feature subset in random. Consider a conditional feature set C containing N features. Let 'p' number of bees be chosen as the population size. From this population, half of the bees are considered as employed bees and the remaining are considered as onlooker bees. For each employed bee, N random numbers are generated between 1 and N and assigned. From these random numbers, the feature subset is constructed by performing round operation and then extracting only the unique numbers from the set. For example, consider the random numbers { 1.45, 1.76, 3.33, 1.01 }, where N=4, first we perform round operation, then the set is modified as { 1, 1, 3 ,1 }. From the above result, the unique numbers alone are extracted as { 1, 3 } and they represent the feature subset. ie., the 1st and 3rd features alone are chosen. In the second step of the algorithm, for each employed bee, whose total number equals to half of the number of food sources, a new source is produced by:

$$v_{ij} = x_{ij} + \varphi_{ij}(x_{ij} - x_{kj}) \qquad (19)$$

where $\varphi_{ij}$ is a uniformly distributed real random number within the range [-1,1], k is the index of the solution chosen randomly from the colony (k = int (rand * N) + 1), j = 1, . . .,D and D is the dimension of the problem. After producing vi, this new solution is compared to $x_i$th solution and the employed bee exploits the better source. In the third step of the algorithm, an onlooker bee chooses a food source with a higher probability and produces a new source in selected food source site. As for the employed bee, the better source is decided to be exploited.

The indiscernibility relation is calculated for each feature subset as objective value ($f_i$). This value has to be maximized. From this objective value the fitness value is calculated for each bee, as given in the following equation:

$$fit_i = \begin{cases} 1/(1+f_i) & if\ f_i \geq 0 \\ 1+abs(f_i) & otherwise \end{cases} \quad (20)$$

The probability is calculated by means of fitness value using the following equation.

$$P_i = \frac{fit_i}{\sum_{j=1}^{N} fit_j} \quad (21)$$

where $fit_i$ is the fitness of the solution xi. After all onlookers are distributed to the sources, sources are checked whether they are to be abandoned. If the number of cycles that a source can not be improved is greater than a predetermined limit, the source is considered to be exhausted. The employed bee associated with the exhausted source becomes a scout and makes a random search in problem domain by the following equation.

$$x_{ij} = x_j^{min} + (x_j^{max} - x_j^{min}) * rand \quad (22)$$

The pseudocode of our proposed method is given as:

ROUGHBEE (C,D)
C, the set of all conditional features;
D, the set of decision features.
 (1) Select the initial parameter values for BCO
 (2) Initialize the population ($x_i$)
 (3) Calculate the objective and fitness value
 (4) Find the optimum feature subset as global.
 (5) do
   a. Produce new feature subset ($v_i$)
   b. Apply the greedy selection between $x_i$ and $v_i$
   c. Calculate the fitness and probability values
   d. Produce the solutions for onlookers
   e. Apply the greedy selection for onlookers
   f. Determine the abandoned solution and scouts
   g. Calculate the cycle best feature subset
   h. Memorize the best optimum feature subset
 (6) repeat        // for maximum number of cycles

Figure 4 Bee Colony based Reduct Algorithm

The parameters used in the proposed method:
| | |
|---|---|
| The population size (number of bees) | 10 |
| The dimension of the population | N |
| Lower bound | 1 |
| Upper bound | N |
| Maximum number of iterations | 1000 |
| The number of runs | 3 |

## 4. EXPERIMENTS & RESULTS

The performance of the reduct calculation approaches discussed in this paper has been tested with 5 different medical datasets obtained from UCI machine learning data repository. Table 1 shows the details of the datasets used in this paper.

TABLE 1
DATASETS USED FOR REDUCT

| Dataset Name | Total Number of Instances | Total Number of Features |
|---|---|---|
| Dermatology | 366 | 34 |
| Cleveland Heart | 300 | 13 |
| HIV | 500 | 21 |
| Lung Cancer | 32 | 56 |
| Wisconsin | 699 | 09 |

TABLE 2
REDUCTS FOUND FOR THE DATASETS

| Dataset | Dermatology | Cleveland Heart | HIV | Lung Cancer | Wisconsin |
|---|---|---|---|---|---|
| #Features | 34 | 13 | 21 | 56 | 09 |
| RSAR | 10 | 7 | 13 | 4 | 5 |
| EBR | 10 | 7 | 13 | 4 | 5 |
| AntRSAR | 8-9 | 6-7 | 10-11 | 4 | 5 |
| GenRSAR | 10-11 | 6-7 | 11-13 | 6-7 | 5 |
| PSORSAR | 7-8 | 6-7 | 9-10 | 4 | 4-5 |
| BeeRSAR | 7 | 6 | 8 | 4 | 4 |

Table 2 shows the reduct results of the various methods, on the 5 different medical datasets. It shows the size of the reduct found for each method. The QuickReduct and EBR methods produced the same reduct every time, unlike GenRSAR, AntRSAR, PSORSAR and BeeRSAR which found different reducts and sometimes different reduct cardinalities. On the whole, it appears to be the case that BeeRSAR outperforms the other methods. But compared to the other methods, BeeRSAR consumes more time to find the reduct.

## 5. CONCLUSION

Feature Selection is an important research direction of rough set application. However, this technique often fails to find better reducts. This paper starts with the fundamental concepts of rough set theory and explains two basic techniques: QuickReduct and Entropy-Based Reduct. These methods can produce close to the minimal reduct set, but not optimal. The swarm intelligence methods have been used to guide this method to find the minimal reducts. Here we have discussed three different computational intelligence based reducts: GenRSAR, AntRSAR and PSO-RSAR. Though these methods are performing well, there is no consistency since they are dealing with more random parameters. In this paper, we have proposed a Bee Colony Optimization algorithm hybrid with Rough set theory to find minimal reducts. This method does not require any random parameter assumption. All these methods are analyzed using medical datasets. As shown in the results, our proposed method exhibits consistent and better performance than the other methods. Here we have employed BeeRSAR approach only for the datasets with numerical attributes. In future, the same approach can be extended to categorical attributes and also to handle missing values.

## REFERENCES

[1] Alpigini, J.J. Peters, J.F. Skowronek, J. Zhong, N. (2002) 'Rough Sets and Current Trends in Computing', Third International Conference, RSCTC 2002, Malvern, PA, USA.

[2] Bonabeau, E. Dorigo, M. and Theraulez, G. (1999) Swarm Intelligence: From Natural to Artificial Systems, Oxford University Press Inc., New York, NY, USA.

[3] Chouchoulas, A. and Shen, Q. (2001) 'Rough set-aided keyword reduction for text categorization', Applied Artificial Intelligence, Vol. 15, No. 9, pp. 843-873.

[4] Dash, M. and Liu, H. (1997) 'Feature Selection for Classification', Intelligent Data Analysis, Vol. 1, No. 3, pp. 131-156.

[5] Holland, J. (1975) Adaptation in Natural and Artificial Systems, The University of Michigan Press, Ann Arbour.

[6] Jensen, R. and Shen, Q. (2001) 'A Rough Set-Aided System for Sorting WWW Bookmarks', In N. Zhong et al. (Eds.), Web Intelligence: Research and Development, pp. 95-105.

[7] Jensen, R. and Shen, Q. (2003) 'Finding rough set reducts with ant colony optimization', Proceedings UK Workshop on Computational Intelligence, pp. 15–22.

[8] Kudo, M. and Skalansky, J. (2000) 'Comparison of algorithms that select features for pattern classifiers', Pattern Recognition, Vol. 33, No. 1, pp. 25-41.

[9] Karaboga, D. (2005) 'An idea based on honey bee swarm for numerical optimization', Technical Report TR06, Erciyes University, Engineering Faculty, Computer Engineering Department.

[10] Karaboga, D. and Basturk, B (2006) 'An Artificial Bee Colony (ABC) algorithm for numeric function optimization', In IEEE Swarm Intelligence Symposium 2006, Indiana, USA.

[11] Karaboga, D. and Basturk, B. (2007a) 'A powerful and efficient algorithm for numerical function optimization: Artificial Bee Colony (ABC) algorithm', Journal of Global Optimization, Vol. 39, No. 3, pp. 459–471.

[12] Karaboga, D. and Basturk, B. (2007b) 'Artificial Bee Colony (ABC) Optimization Algorithm for Solving Constrained Optimization Problems' Foundations of Fuzzy Logic and Soft Computing', LNCS, Springer-Verlag, Vol. 4529, pp. 789–798.

[13] Karaboga, D. and Basturk, B. (2008) 'On the performance of Artificial Bee Colony (ABC) algorithm', Applied Soft Computing, Vol. 8, No. 1, pp. 687–697.

[14] Liu, H. Abraham, A. and Li, Y. (2009) 'Nature Inspired Population-Based Heuristics for Rough Set Reduction', Rough Set Theory, SCI, Springer-Verlag, Vol. 174, pp. 261-278.

[15] Pawlak, Z. (1982) 'Rough Sets', International Journal of Computer and Information Sciences, Vol. 11, pp. 341–356.

[16] Pawlak, Z. (1991) Rough Sets: Theoretical Aspects of Reasoning about Data, Kluwer Academic Publishers.

[17] Pawlak, Z. (1993) 'Rough Sets: Present State and The Future', Foundations of Computing and Decision Sciences, Vol. 18, pp. 157–166.

[18] Pawlak, Z. (2002) 'Rough Sets and Intelligent Data Analysis', Information Sciences, Vol. 147, pp. 1–12.

[19] Quinlan, J.R. (1993) C4.5: Programs for Machine Learning, The Morgan Kaufmann Series in Machine Learning. Morgan Kaufmann Publishers, San Mateo, CA.

**Author Biographies**

**N.Suguna** received her B.E degree in Computer Science and Engineering from Madurai Kamaraj University in 1999 and M.E. degree in Computer Science and Engineering from Bangalore University in 2003. She has more than a decade of teaching experience in various Engineering colleges in Tamil Nadu and Karnataka. She is currently with Akshaya College of Engineering and Technology, Coimbatore, Tamilnadu, India. Her research interests include Data Mining, Soft Computing and Object Oriented Systems.

**Dr.K.Thanushkodi** has more than 35 years of teaching experience in various Government & Private Engineering Colleges. He has published 45 papers in International journals and conferences. He is currently guiding 15 research scholars in the area of Power System Engineering, Power Electronics and Computer Networks. He has been the Principal in-charge and Dean in Government College of Engineering Bargur. He has served as senate member in Periyar University, Salem. He has served as member of the research board, Anna University, Chennai. He Served as Member Academic Council, Anna University, Chennai. He is serving as Member, Board of Studies in Electrical Engineering, Anna University, Chennai. He is serving as Member, Board of Studies in Electrical and Electronics & Electronics and Communication Engineering, Amritha Viswa Vidya Peetham, Deemed University, Coimbatore. He is serving as Governing Council Member SACS MAVMM Engineering College, Madurai. He served as Professor and Head of E&I, EEE, CSE & IT Departments at Government College of Technology, Coimbatore. He is serving as Syndicate Member of Anna University, Coimbatore. Currently, he is the Director of Akshaya College of Engineering and Technology, Coimbatore.